\documentclass{article}

\usepackage{arxiv}

\usepackage[utf8]{inputenc} 
\usepackage[T1]{fontenc}    
\usepackage{hyperref}       
\usepackage{url}            
\usepackage{amsfonts}       
\usepackage{nicefrac}       
\usepackage{microtype}      
\usepackage{lipsum}
\usepackage{graphicx}
\usepackage{amsmath}
\usepackage{bbm}
\usepackage{booktabs}
\usepackage{wrapfig}
\usepackage{multirow}
\usepackage[capitalise]{cleveref}
\usepackage{algorithm,algpseudocode}
\usepackage[algo2e]{algorithm2e} 
\usepackage{amssymb}
\usepackage{amsfonts}
\usepackage{diagbox}
\usepackage{color}
\usepackage{bm}
\usepackage{adjustbox}
\usepackage[numbers]{natbib}
\usepackage{tikz}
\usepackage{caption}
\usetikzlibrary{bayesnet}
\usetikzlibrary{arrows}
\usetikzlibrary{backgrounds}
\graphicspath{ {./images/} }
\newtheorem{definition}{Definition}

\title{Bias-Tolerant Fair Classification}

\author{
 Yixuan Zhang \\
  Data Science Institute\\
  University of Technology Sydney\\
  \texttt{yixuan.zhang@student.edu.au} \\
   \And
 Feng Zhou \\
  Department of Computer Science and Technology\\
  Tsinghua University\\
  \texttt{zhoufeng6288@tsinghua.edu.cn} \\
  \And
 Zhidong Li \\
  Data Science Institute\\
  University of Technology Sydney\\
  \texttt{zhidong.li@uts.edu.au} \\
    \And
 Yang Wang\\
  Data Science Institute\\
  University of Technology Sydney\\
  \texttt{yang.wang@uts.edu.au} \\
    \And
 Fang Chen \\
  Data Science Institute\\
  University of Technology Sydney\\
  \texttt{fang.chen@uts.edu.au} \\
}

\begin{document}
\maketitle
\begin{abstract}
The label bias and selection bias are acknowledged as two reasons in data that will hinder the fairness of machine-learning outcomes. The label bias occurs when the labeling decision is disturbed by sensitive features, while the selection bias occurs when subjective bias exists during the data sampling. Even worse, models trained on such data can inherit or even intensify the discrimination. Most algorithmic fairness approaches perform an empirical risk minimization with predefined fairness constraints, which tends to trade-off accuracy for fairness. However, such methods would achieve the desired fairness level with the sacrifice of the benefits (receive positive outcomes) for individuals affected by the bias. Therefore, we propose a \textbf{B}ias-Tolerant \textbf{FA}ir \textbf{R}egularized \textbf{L}oss (B-FARL), which tries to regain the benefits using data affected by label bias and selection bias. B-FARL takes the biased data as input, calls a model that approximates the one trained with fair but latent data, and thus prevents discrimination without constraints required. In addition, we show the effective components by decomposing B-FARL, and we utilize the meta-learning framework for the B-FARL optimization. The experimental results on real-world datasets show that our method is empirically effective in improving fairness towards the direction of true but latent labels.
\end{abstract}


\section{Introduction}
\label{introduction}
With the increasing adoption of autonomous decision-making systems in practice, the fairness of the outcome obtained from such systems has raised widespread concerns~\cite{fair_transfer,fair_constraints}. As the decision-making systems are driven by data and models, they are vulnerable to data bias since the model can replicate the biases contained in the input data and output biased decisions~\cite{bird2016exploring}. To address the issues, researchers proposed many fairness-aware learning methods and demonstrated the potential in dealing with discrimination problems in job applicants selection~\cite{company_example}, credit card approval~\cite{credit_example} and recidivism prediction~\cite{compas}. The fairness-aware learning methods in the previous work can be categorized into (1) pre-processing methods: learn fair representations of the input data~\cite{2013_iclr_ae,icml_2013,NIPS2017_optimised_preprocessing,2016_lum}; (2) in-processing methods: incorporate fairness constraints into the objective function to achieve certain level of fairness~\cite{fair_constraints,2016_zafar,5360534,2018_icml_reductions,Kamishima_2011} and (3) post-processing methods~\cite{equal_opportunity}: modify the learned posterior distribution of the prediction to achieve fairness. In this paper, we mainly focus on the second category, where the approaches perform an empirical risk minimization with predefined fairness constraints. These constraints, heavily dependent on predefined fairness definitions, are combined with the loss to be a fairness-aware objective function.


Model optimization based on the fairness-aware objective function creates the controversy of the trade-off between accuracy and fairness~\cite{berk2017convex}. The recent work of~\cite{unlocking_fairness} presented the paradox that accuracy drops due to the ignorance of \emph{label bias} and \emph{selection bias} when imposing fairness constraints to the model. By definition, the label bias will flip the label, e.g., from `qualified' to `unqualified' in recruitment data; and the selection bias will distort the ratios between the protected and unprotected group, e.g., select less positive labeled instances from the protected group. The reason that trade-off occurs is that the accuracy is still evaluated on the biased data. However, when evaluated on the bias-free data, fairness and accuracy should improve simultaneously. 

In this work, inspired by the peer loss~\cite{liu2020peer}, we propose the loss function, B-FARL, that can automatically compensate both selection bias and label bias existing in input data with implicit regularizers. By minimizing the loss, the learned classifier using biased data is equivalent to the learned one using unbiased data. The peer loss is designed to handle binary label noise problems where labels are flipped randomly conditioning on the true class. It is similar to the label bias setting in our problem but has no dependence between the flip rate and sensitive features. In the design of our B-FARL, the flip rate is separately considered for distinct demographic groups (samples with different values of sensitive feature). B-FARL inherits the strength of peer loss which does not require flip rate estimation; in addition, B-FARL also does not require explicit fairness constraints or the level of fairness violation. We will show and prove that B-FARL is an appropriate loss function that guides the model to learn towards fair prediction from the biased data. 

Furthermore, though peer loss does not require noise rate estimation, it requires tuning a noise rate related hyperparameter via cross validation, which is time consuming. To address this issue, we utilize the meta-learning framework. Meta-learning can learn meta-parameters (parameters to be optimized) from data directly, which is a data-driven optimization framework. Motivated by the success of hyperparameter optimization using meta-learning~\cite{meta_learning}, we incorporate our B-FARL into the model-agnostic meta-learning (MAML) optimization framework to dynamically update the hyperparameters, which is more efficient than cross validation. 

Specifically, our work makes three main contributions: \textbf{(1)} We propose the B-FARL, which enables the learning of a fair model using data containing label bias and selection bias. It is worth nothing that B-FARL does not require predefined fairness constratins but learns fairness directly from data. \textbf{(2)} We provide a theoretical analysis of the effectiveness of B-FARL by decomposing it into three indicative terms, i.e., the expected loss on the distribution of clean data, a fairness regularizer w.r.t. subgroups risk deviation, and the regularizer on the disagreement between biased and unbiased observations. \textbf{(3)} We utilize MAML framework to optimize the noise rate related hyperparameters, which is more efficient than the traditional cross validation.



\section{Related Work}
\label{related_work}
\paragraph{Fairness in machine learning}
Most algorithmic fairness approaches in the literature incorporate fairness constraints into the objective function~\cite{fair_constraints,2016_zafar,5360534,2018_icml_reductions,Kamishima_2011} for optimization. The fairness constraints need to be predefined according to various statistical fairness criteria, such as equality opportunity~\cite{equal_opportunity}, equalized odds~\cite{equal_opportunity} and demographic parity notion like p\%-rule~\cite{biddle}. In the work of ~\cite{donini2020empirical} and ~\cite{Rezaei_2020}, they proposed to use the nonlinear measure of dependence as regularizers to approximate p\%-rule or equality opportunity violations. However, the approximation could potentially hurt the performance. Besides, there are two main general drawbacks to these methods. First, the fairness criteria must be carefully chosen. Second, if the constraints can grant a fair model, testing it on the biased data will hurt the accuracy. This creates the controversy of the trade-off between accuracy and fairness. The recent work of~\cite{unlocking_fairness} analyzed the second drawback by a framework that considered label bias and selection bias. Under the bias setting, deploying fairness constraints directly to the biased data can both hurt the accuracy and fairness. To address the issue, we propose to incorporate algorithmic fairness by the label noise framework that can handle biased data learning. The most similar work is~\cite{gpl}. However, this work is fundamentally different from ours w.r.t. the problem to be solved. Their problem is how to derive fairness constraints on corrupted data in the label noise problem, while we solve the fairness problem by considering the label bias and selection bias as a special type of label noise.



\paragraph{Noisy label learning}
Most recent works of learning from noisy labels focus on modifying the loss function, which include loss correction and reweighting methods~\cite{Scott,NIPS2013_5073,importance_reweighting,patrini2017making}. However, these methods require estimating the noise rate or cannot handle asymmetric noise rates. The recent work of~\cite{liu2020peer} proposed a peer loss function based on the idea of peer prediction to solve label noise problems under the asymmetric noise setting. The peer loss function is defined as subtracting the loss of random sampled feature-label pair from the loss of each sample. This method does not require noise rate and enables us to perform empirical risk minimization on corrupted data. The loss proposed in our work is related to the $\text{CORES}^2$ (COnfidence REgularized Sample Sieve)~\cite{cheng2020learning} that improves the performance of peer loss by taking the expectation of the robust cross-entropy loss over the random sample pairs, encouraging a more confident prediction. This work inspires us to propose the B-FARL to solve the discrimination problem from a label bias perspective. However, this work does not in an end-to-end manner, it separates the learning process into two phases: select most clean samples in the first phase and treats the rest samples as unlabeled and retrain the model in the second phase.

\section{Proposed Method}
\label{problem_formulation}
In this section, we will present our design for B-FARL. We begin with a detailed problem formulation. Next, we introduce the methodology of B-FARL followed by the analysis of B-FARL. At last, we provide the algorithm for optimizing B-FARL.

\subsection{Problem Formulation}
\begin{wrapfigure}{R}{0.3\textwidth}
\vspace{.15in}
\hspace{.15in}
\begin{tikzpicture}[]
\centering
\captionsetup{justification   = raggedright,
              singlelinecheck = false}
    \node[latent] (w) {$\mathbf{W}$} ; %
    \node[latent, right=of w] (z) {$Z$} ; %
    \node[obs, right=of z] (y) {$Y$} ; %
    \node[obs, above=of z] (x) {$X$} ; %
    \node[obs, above=of y] (a) {$A$} ; %
    \plate[inner sep=0.2cm, xshift=-0.02cm, yshift=0.12cm] {plate2} {(w) (z) (a) (x) (y)} {N}; %
    \edge {w} {z} ; %
    \edge {z} {y} ; %
    \edge {x} {z} ; %
    \edge {a} {y} ;
\end{tikzpicture}
\caption{Generative process of bias in $N$ observations, shaded nodes are observations.}
\label{fig:generated}
\end{wrapfigure}
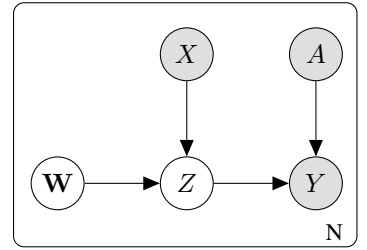
Given the triplet of random variables $(X,Z,A)$ with sample space $\Omega = \mathcal{X} \times \{-1,1\} \times \{0,1\}$, $X$ denotes the non-sensitive feature, $Z$ denotes the clean and fair label and $A$ is the binary sensitive feature. Let $f:X\rightarrow Z$ be a fair labeling function, which maps $X$ to a fair and clean outcome $Z$. To obtain observations, we can use an observation distribution $D$ to generate samples for the triplet. When the generative process is independent of $A$, we name $D$ clean and fair distribution since the data will be fair. However, in our problem, we assume $D$ and the generated data are latent because of discrimination. In the framework proposed by~\cite{unlocking_fairness}, we can decompose the discrimination as label bias and selection bias. So, instead of observing samples from the true distribution $D$, we assume one can only observe samples from a corrupted distribution $\widetilde{D}$, where the labels from $\widetilde{D}$ are discriminated by sensitive feature $A$. We denote the discriminated label as $Y$ and we assume $Z$ is flipped to $Y$ with the probability conditioning on $A$, i.e., $\theta^{\text{sgn}(y)}_a = P(Y=y\mid Z=-y,A=a)$ in the binary classification setting. We summarize the process of labels being discriminated in Fig.~\ref{fig:generated}. We also assume $A$ is independent of $X$. Such a setting separates the discrimination from features and lets all the sources of discrimination be in $A$.

The label bias is from biased decisions on the sensitive feature, e.g., gender or race. Label bias can cause the function $f$ learned from $(X,Y,A)$ being discriminated. On the other side, different from label bias, selection bias will affect the true ratio of two demographic groups in favor of positive outcome ($Z=1$), and affect the data distribution $D$ in further. We assume the selection bias occurs in the process of selecting samples from positive labeled instances among the protected group and we denote the selection bias as $\frac{r}{\sigma}$, where $r$ is the original proportion of positive labeled instances among the protected group and $\sigma=1$ if no selection bias occurs while $\sigma >1$ if selection bias occurs. The selected data is denoted as $\hat{D}$ which is a subset of $D$. Our aim is to learn a labeling function $\hat{f}$ under the corrupted distribution $\widetilde{D}$ that can approximate the fair labeling function $f$ and hence enable the prediction toward fairness. We propose to use noisy label learning methods to solve this problem. Some of these techniques,  such as the re-weighting~\cite{NIPS2013_5073,importance_reweighting} or loss correction~\cite{patrini2017making} methods, require $\theta$ to be known, or they cannot handle asymmetric noise rates. To be more robust, we will eliminate such a requirement by addressing it with peer loss~\cite{liu2020peer}. 

A noticeable challenge of the solution is that only label bias is convertible to the label noise, while selection bias and the combined bias cannot be directly fit into it. With the assumption that the selection bias occurs in the process of selecting positive labeled instances among the protected group, it will affect $\theta_0^-$. Let $\varepsilon_0^{-}$ denote the bias rate combining the selection bias and label bias to represent the proportion that how many data among protected group labeled as $+$ are finally observed as $-$.  The relationship between $\varepsilon_0^-$ and $\theta_0^-$ can be derived as $\theta_0^- = \frac{\sigma-r}{1-r}\varepsilon_0^- + \frac{1-\sigma}{1-r}$. The full derivation can be found in the Appendix \ref{apd:second}.

\subsection{B-FARL}
In this section, we present our design for B-FARL based on peer loss. For each sample $(x_i,y_i)$, the peer loss~\cite{liu2020peer} for $i$ is defined as
\begin{equation}
\begin{aligned}
\ell_{peer} = \ell (f(x_i,\bm{\omega}),y_i) - \alpha \cdot \ell (f(x_{i_1},\bm{\omega}),y_{i_2}),
\end{aligned}
\label{eq:peer loss}
\end{equation}
where $\alpha$ is used as the parameter to make peer loss robust to unbalanced labels, and computed as
\begin{equation}
{\small
\begin{aligned}
\alpha := 1-(1-P(Y=-1\mid Z=+1)-P(Y=+1\mid Z=-1)) \frac{P(Z=+1)-P(Z=-1)}{P(Y=+1)-P(Y=-1)}.
\end{aligned}}
\end{equation}
In other words, when $P(Z=+1) = P(Z=-1) = 0.5$, $\alpha$ is $1$. In practice,  $\alpha$ can be tuned as a hyperparameter~\cite{liu2020peer}, which means we do not require to know $P(Z=+1)$ and $P(Z=-1)$ for computing $\alpha$. In~\cref{eq:peer loss}, $i_1,i_2$  are independently sampled from $S/\{i\}$ ($S = \{1,2,\dots,N\}$) by $\frac{1}{N}$. The corresponding random variables with sensitive attribute are the triplet of $(X_{i_1},A_{i_1},Y_{i_2})$.

If we take demographic groups into consideration, the original peer loss is re-weighted by a factor $\delta_a$. Similar to~\cite{gpl}, it is defined as $\delta_a = \frac{1}{1-\theta_a^+-\theta_a^-}$ and hence the group-weighted peer loss for $i$ is 
\begin{equation}
\begin{aligned}
\ell_{gp} =\delta_{a_i}\cdot \ell_{peer}.
\end{aligned}
\end{equation}
According to~\cite{liu2020peer}, $\delta_a$  used to re-scale peer loss on biased data to clean data. Then we will show how B-FARL is designed by decomposing $\ell_{gp}$ for the protected and unprotected groups. First, we take the expectation of $\ell_{gp}$ w.r.t. $X_{i_1}$ and $Y_{i_2}$ over distribution conditioning on $A$ as \cref{eq:derive}. There are two other reasons to take the expectation form: (1) the expectation form enables us to write the loss in terms of $x_i$ rather than the random variable $X_{i_1}$, which provides convenience for computing. (2) instead of randomly sampled pairs, we use the expectation to keep the loss stable.


\begin{equation}
\label{eq:derive}
{\small 
    \begin{aligned}
    & \frac{1}{N}\sum_{i=1}^N\mathbb{E}_{X_{i_1},Y_{i_2}\mid \widetilde{D}}[\delta_{a_i}(\ell(f(x_i,\bm{\omega}),y_i) - \alpha \cdot \ell (f(X_{i_1},\bm{\omega}),Y_{i_2}))] \\ 
    & = \frac{1}{N}\sum_{i}^N\delta_{a_i}[\ell(f(x_i,\bm{\omega}),y_i) - \alpha \cdot P(A=0\mid \widetilde{D})\sum_{i'\in S_0}P(X_{i_1}=x_{i'}\mid A=0, \widetilde{D})\mathbb{E}_{Y\mid \widetilde{D},A=0} \ell (f(x_{i'},\bm{\omega}),Y) \\
    & - \alpha \cdot P(A=1\mid \widetilde{D})\sum_{i'\in S_1}P(X_{i_1}=x_{i'}\mid A=1, \widetilde{D})\mathbb{E}_{Y\mid \widetilde{D},A=1} \ell (f(x_{i'},\bm{\omega}),Y)] \\
     &= \frac{1}{N}\sum_{i}^N\delta_{a_i}[\ell(f(x_i,\bm{\omega}),y_i) - 
    \alpha \cdot \frac{|S_0|}{N}\sum_{i'\in S_0}\frac{1}{|S_0|}\mathbb{E}_{Y\mid \widetilde{D},A=0} \ell (f(x_{i'},\bm{\omega}),Y)\\
    &- \alpha \cdot \frac{|S_1|}{N}\sum_{i'\in S_1}\frac{1}{|S_1|}\mathbb{E}_{Y\mid \widetilde{D},A=1} \ell (f(x_{i'},\bm{\omega}),Y)] \\ 
    &= \frac{1}{N}(\sum_{i\in S_0}\delta_{a_i}[\ell(f(x_i,\bm{\omega}),y_i) - \alpha \cdot \mathbb{E}_{Y\mid \widetilde{D},A=0}\ell(f(x_i,\bm{\omega}),Y)] + \sum_{i\in S_1}\delta_{a_i}[\ell(f(x_i,\bm{\omega}),y_i) - \alpha \cdot \mathbb{E}_{Y\mid \widetilde{D},A=1}\ell(f(x_i,\bm{\omega}),Y)]),
    \end{aligned}}
\end{equation}
where $S_0 = \{i|a_i=0\}$ and $S_1 = \{i|a_i=1\}$. Based on~\cref{eq:derive}, we add intensity parameter to obtain the framework of B-FARL (${L_F}$) as

\begin{equation}
L_F= \frac{1}{N}\sum^N_{i=1}(\ell_B(\bm{\omega}) + \bm{\beta} \bm{\ell}_{A}(\bm{\omega})),
\label{eq:new_loss}
\end{equation}
with
\begin{equation}
\begin{gathered}
    \ell_B(\bm{\omega}) = \delta_{a_i}\ell(f(x_i,\bm{\omega}),y_i),\;\;\; \bm{\beta} = \begin{bmatrix}
    -\beta_0 \\ -\beta_1
    \end{bmatrix}^T,\\
    \bm{\ell}_{A}(\bm{\omega}) = 
    \begin{bmatrix}
   \mathbb{E}_{Y\mid  \widetilde{D},A=0} (1-a_i)\ell(f(x_i,\bm{w}),Y) \\
   \mathbb{E}_{Y\mid \widetilde{D},A=1}a_i \ell(f(x_i,\bm{w}),Y)
    \end{bmatrix},
\end{gathered}
\end{equation}
where $\beta_0$, $\beta_1$ are two hyperparameters that control the intensity of the regularizer terms~($\bm{\ell}_{A}$). We let 
$\delta_{a_i}$ and $\alpha$ in \cref{eq:derive} be absorbed into $\beta_0$ and $\beta_1$. Most widely used surrogate loss functions can be used for $\ell$. For example, 0-1 loss can be applied with sufficient training data~\cite{bartlett2006convexity} for its robustness to instance-dependent noise~\cite{Manwani_2013} but alternatives also can be applied such as cross entropy, logistic loss, etc. Compared to the peer loss, the two expectation regularization terms conditioning on the protected and non-protected groups can further improve the prediction performance. In section~\ref{analysis}, we will show how the regularization terms help improve the performance.



\subsection{Analysis of the B-FARL}
\label{analysis}
In this section, we explain the effectiveness of \cref{eq:new_loss} by decomposing it into components that demonstrate fairness regularization and discrimination correction. The full derivation can be found in Appendix~\ref{apd:decompose}. B-FARL can be decomposed into the following three terms


\begin{equation}
\label{eq:analysis}
    \begin{aligned}
    & \mathbb{E}_{\widetilde{D}}[\ell_B(\bm{\omega}) + \bm{\beta} \bm{\ell}_{A}(\bm{\omega})]\\
    = & \underbrace{\mathbb{E}_{D}[\ell(f(X),Z)]}_\text{clean model}
    + \underbrace{\lambda \cdot |\mathbb{E}_{\widetilde{D}\mid A=0}\ell(f(X),Y) - \mathbb{E}_{\widetilde{D}\mid A=1}\ell(f(X),Y)|}_\text{fairness regularization}\\
    &
    + \underbrace{\sum_a P(A=a) \sum_{k \in [C]}\sum_{l\in [C]}P(Z=l)\mathbb{E}_{D_{x \mid l,a}}(U_{lk}(x,a)-\gamma_a\cdot P(Y=k))\ell (f(x),k).}_\text{bias regularization}
    \end{aligned}
\end{equation}
The first term is for learning with clean data. The second term shows the fairness regularization w.r.t. subgroup risks deviation which is defined in Def.~\ref{def:definition}. The last term shows the regularization effect on the biased data. Here both the regularizer effects $\lambda$ in the second term and $\gamma_a$ in the last term are decomposed from $\beta_0$ and $\beta_1$ in \cref{eq:new_loss}.

\begin{definition}[Perfect fairness via subgroup risks]
We say that a predictor $f \in \mathcal{F}$ is perfectly fair w.r.t. a loss function $\ell$ if all subgroups attain the same average loss; i.e., in the binary sensitive attributes case (Sec. 3.2 in ~\cite{pmlr-v97-williamson19a}),
\begin{equation}
    \mathbb{E}_{X,Y\mid A=0}\ell(f(X),Y) = \mathbb{E}_{X,Y\mid A=1}\ell(f(X),Y).
\end{equation}
\label{def:definition}
\end{definition}
More specifically:
\begin{itemize}
\item The first term is the expected loss on the distribution of clean samples.

\item The second term is a fairness regularizer on the noisy distribution w.r.t. the subgroup risk measure on the noisy distribution. As explained in~\cite{pmlr-v97-williamson19a}, Def.~\ref{def:definition} tells us under the perfect fairness, the prediction performance w.r.t. the sensitive attributes should not vary. The best case for the regularizer is perfect fairness according to Def.~\ref{def:definition}. We use the difference between average subgroup risk to measure the fairness violation and $\lambda$ is the regularizer effect. 

\item The third term is a regularizer w.r.t. noisy loss. This loss is the penalty for the disagreement between $Y$ and $Z$. The ideal situation is that $(U_{lk}(x,a)-\gamma_a\cdot P(Y=k))$ should be minimized, where $U_{lk}(x,a) = \begin{cases}
        \delta_a \theta_a^{\text{sgn}(k)} \text{     if $l \neq k$,}\\
        \delta_a \theta_a^{\text{sgn}(l)} \text{    if $l=k$}.
        \end{cases}$, and hence the noisy term will vanish. We should point out that the selection bias is included in $\theta_1^- = \frac{\sigma -r}{1-r}\varepsilon_1^- + \frac{1-\sigma}{1-r}$ and if $\sigma=1$, $\theta_1^- = \varepsilon_1^-$.
\item For equivalence, it is noticeable when the first term is minimized, $f(X)$ is the Bayes optimal classifier on clean data, which means the penalties of all bias do not exist. As a result, on the optimal point, all three terms are minimized so that the summation is also minimized. Therefore, classifier that can minimize the B-FARL equals classifier that can minimize the first term, which indicates the equivalence.
\item The effectiveness of the first and second terms are similar to traditional loss function with fairness constraints. However, here the loss function is learned from $Z$ while the traditional methods still use $Y$. Such difference endues our loss the capability to learn the correct model. 
\end{itemize}

\subsection{Optimization B-FARL via Model-Agnostic Meta-Learning}
\label{model_setup}
Meta-learning is a general framework of data-driven optimization. Most of the meta-learning methods can be viewed as a bi-level optimization which contains inner loop optimization (main optimization) and outer loop optimization (optimize the meta-parameter, e.g. hyperparmeters of inner loop). In our work, we consider the B-FARL as the main optimization goal and the re-weighting factor $\delta_{a_i}$ and regularization parameters $\bm{\beta}$ as the meta-parameters. Since $\delta_{a_i}$ for individuals among the same demographic group is the same, we can also write the first part in \cref{eq:new_loss} as the following format
\begin{equation}
{\small
\begin{aligned}
  \frac{1}{N}\sum^N_{i=1}\ell_B(\bm{\omega})
   & = \frac{1}{N} [\alpha_0\sum_{i\in\{S_0\}} \ell(f(x_i,\bm{\omega}),y_i) + \alpha_1 \sum_{i\in\{S_1\}}\ell(f(x_i,\bm{\omega}),y_i) ]= \frac{1}{N} \bm{\alpha} \bm{\ell}_{D_a},
\end{aligned}
}
\end{equation}
where $\bm{\alpha} = \begin{bmatrix} \alpha_0 \\ \alpha_1 \end{bmatrix}^T$ and $\bm{\ell}_{B_a} = [\sum_{i\in\{S_0\}} \ell(f(x_i,\bm{\omega}),y_i), \sum_{i\in\{S_1\}}\ell(f(x_i,\bm{\omega}),y_i)]$. Overall, the optimization can be viewed as
\begin{equation}
    \min_{\bm{\alpha},\bm{\beta}} L_F(\bm{\omega_p}), \bm{\omega_p} = \arg \min_{\bm{\omega}} L_F(\bm{\omega}).
\end{equation}
We split the optimization into two stages and here we define $\bm{\omega}^t$, $\bm{\beta}^t$ and $\bm{\alpha}^t$ as the corresponding variables in step $t$. In the meta training stage, we first initialize $\bm{\beta}$ and $\bm{\alpha}$, to obtain $\bm{\omega}^{1}$, then fix $\bm{\omega}^{1}$ to obtain $\bm{\beta}^{1}$ and $\bm{\alpha}^{1}$. These two steps iteratively used to obtain $\bm{\omega}^{t+1}$, $\bm{\beta}^{t+1}$ and $\bm{\alpha}^{t+1}$. In the actual training stage, we optimize B-FARL with the updated $\bm{\beta}^{t+1}$ and $\bm{\alpha}^{t+1}$ from meta training stage. The detailed steps are summarized in Algorithm~\ref{alg:optimization}.

\subsubsection{Meta training stage}
We randomly split the training set into mini-batches with batch size $n$. With fixed values of $\bm{\beta}^{t+1}$ and $\bm{\alpha}^{t+1}$, we first perform the inner loop optimization and the one-step-forward weights $\bm{\omega}^{t+1}$ is updated by gradient descent with learning rate $\eta$
\begin{equation}
\label{eq:meta1}
    \bm{\omega}^{t+1} = \bm{\omega}^{t} - \eta\nabla_{\bm{\omega}^t}\frac{1}{n}\sum^n_{i=1}(
    \bm{\alpha}^t \bm{\ell}_{B_a}(\bm{\omega}^t) + {\bm{\beta}}^t \ell_{A}(\bm{\omega}^t))
\end{equation}
Now with updated $\bm{\omega}^{t+1}$, we then perform the outer loop optimization which updates $\bm{\beta}^{t+1}$ and $\bm{\alpha}^{t+1}$ via gradient descent with learning rate $\eta'$
\begin{equation}
\label{eq:meta2}
\begin{gathered}
      \bm{\beta}^{t+1} = \bm{\beta}^{t} - \eta'\nabla_{\bm{\beta}^t}\frac{1}{n}\sum^m_{i=1}(\bm{\alpha}^t \bm{\ell}_{B_a}(\bm{\omega}^{t+1}) + {\bm{\beta}^t} \ell_{A}(\bm{\omega}^{t+1})),\\
      \bm{\alpha}^{t+1} = \bm{\alpha}^{t} - \eta'\nabla_{\bm{\alpha}^t}\frac{1}{n}\sum^m_{i=1}(\bm{\alpha}^t \bm{\ell}_{B_a}(\bm{\omega}^{t+1}) + {\bm{\beta}^t} \ell_{A}(\bm{\omega}^{t+1})).
\end{gathered}
\end{equation}

\subsubsection{Actual training stage}
We should point out that in the meta training stage, $\bm{\omega}$ is the auxiliary as the purpose of meta training stage is to determine the optimal value for $\bm{\beta}$ and $\bm{\alpha}$. Once we have updated $\bm{\beta}$ and $\bm{\alpha}$, we train the model ($\bm{\omega}$ in the actual training stage) via gradient descent with learning rate $\gamma$
\begin{equation}
\label{eq:actual_train}
    \bm{\omega}^{t+1} = \bm{\omega}^{t} - \gamma\nabla_{\bm{\omega}^t}\frac{1}{n}\sum^n_{i=1}(\bm{\alpha}^{t+1} \bm{\ell}_{B_a}(\bm{\omega}^t) + \bm{\beta}^{t+1} \ell_{A}(\bm{\omega}^t)).
\end{equation}





\begin{algorithm}[H]
\SetAlgoLined
Initialize the hyperparameter $\bm{\beta}$ and $\bm{\alpha}$ and model weights $\bm{\omega}$ \;

\For{t=1,$\cdots$ T}{
  Update the model parameter $\bm{\omega}^{t+1}$ by \cref{eq:meta1}\;
  
 Update $\bm{\beta}^{t+1}$ and $\bm{\alpha}^{t+1}$ by \cref{eq:meta2} \;
 
  Train model with $\bm{\beta}^{t+1}$ and $\bm{\alpha}^{t+1}$ by \cref{eq:actual_train}
 }
Obtain the prediction results 
\caption{Optimization for B-FARL}
\label{alg:optimization}
\end{algorithm}

\section{Experiments and Comparisons}
\label{experiment}
In this section, we conduct experiments on real world data to investigate the effects of label bias and selection bias that affect accuracy and fairness and show the effectiveness of our proposed method. Since we cannot observe the latent fair labels of the real-world data, we assume the observed data is clean and add different biases to create a biased version. 

\subsection{Experiment Setup}
In this section, we introduce our experiment setting including the evaluation metrics and dataset descriptions.
\subsubsection{Evaluation Metrics}
We use two metrics: Difference of Equal Opportunity (DEO)~\cite{equal_opportunity} and p\%-rule~\cite{biddle} to measure fairness violation . They are defined as
\begin{equation*}
{\small
\begin{aligned}
\text{DEO} &= |P(\hat{Y}=1 \mid A=1, Y=1) - P(\hat{Y}=1 \mid A=0, Y=1)|,\\
\text{p\%} &= \text{min}(\frac{P(\hat{Y}=1\mid A=0)}{P(\hat{Y}=1\mid A=1)}, \frac{P(\hat{Y}=1\mid A=1)}{P(\hat{Y}=1\mid A=0)}).
\end{aligned}}
\end{equation*}
A higher DEO and smaller p\% indicate more fairness violation. These two indicators evaluate fairness from a different perspective. DEO considers the additional condition with the original label is positive, and p\%-rule only considers the prediction results. Their combination can avoid the case that classifier pushes the results to demographic parity but neglect the true labels. In our experiment, we implement a simple Multi-Layer Perceptron (MLP) to train, and we applied binary cross-entropy loss for $\ell$ in \cref{eq:new_loss}. We use the weighted macro F1 score to measure the performance, which is the macro average weighted by the relative portion of samples within different classes. We split the data into 90\% train and 10\% test, and we report the results in the form of mean $\pm$ standard deviation over ten experiments with ten random splits.

\subsubsection{Dataset Description}

\textbf{Adult Dataset\footnote{\url{http://archive.ics.uci.edu/ml/datasets/Adult}}:} The target value is whether an individual's annual income is over \$50k. The original feature dimension for this dataset is 13. After feature aggregation and feature encoding , the feature dimension is 35. The sensitive attribute is `Gender', and we consider `Gender = Female' as protected group.\\
\textbf{German Credit Dataset\footnote{\url{https://archive.ics.uci.edu/ml/datasets/statlog+(german+credit+data)}}:} The task of this dataset is to classify people as good or poor credit risks. The features include economical situation of each individual as well as personal information like age, gender, personal status, etc. The feature dimension is 13. In our experiment, we set `Gender' as sensitive attribute and `Gender = Male' as protected group. \\
\textbf{Compas Dataset\footnote{ \url{www.propublica.org/article/how-we-analyzed-the-compasrecidivism-algorithm}}:} This data is from COMPAS, which is a tool used by judges, probation and prole officers to asses the risk of a criminal to re-offend. We focus on the predictions of `Risk of Recidivism' (Arrest). The algorithm was found to be biased in favor of white defendants over a two-year follow-up period. We consider `Race' to be the sensitive attribute and `Race=Black' as protected group. After feature encoding and aggregation, the feature dimension is 11. \\

\begin{table}[!h]
\centering
\label{tab:data_description}
\resizebox{\linewidth}{!}{%
\begin{tabular}{|c||c|c|c|c|} 
\toprule
dataset & number of instances & protected/unprotected groups & number of instances & Source \\ 
\hline
Adult & 30,717 & female/male & 10,067/20,650 & UCI \\
German Credit& 900 & female/male & 278/622 & UCI \\
Compas & 6,492 & black/white & 3,325/3,167 & COMPAS \\
\bottomrule
\end{tabular}
}
\caption{Dataset description }
\end{table}

\subsubsection{Baseline Models}
From the perspective of fairness constraints, we compare to two recent fairness-aware learning methods:~\cite{Rezaei_2020}; ~\cite{donini2020empirical}; From the perspective of label bias, we compare to two related noisy label learning methods: $\text{CORES}^2$~\cite{cheng2020learning}; Group Peer Loss (GPL)~\cite{gpl} as our baseline comparison. Besides, we also compare to two baseline methods: Clean and Biased, in which we train MLP on the clean data and biased data respectively.

For the effeciency, the runtime of GPL is around 20.51 minutes. B-FARL only needs 0.83 minutes. $\text{CORES}^2$ needs 2.32 minutes for two phases together. The incorporation of the meta-learning framework is much more efficient.

\begin{figure}[htp]
\begin{center}
\includegraphics[width=1\textwidth]{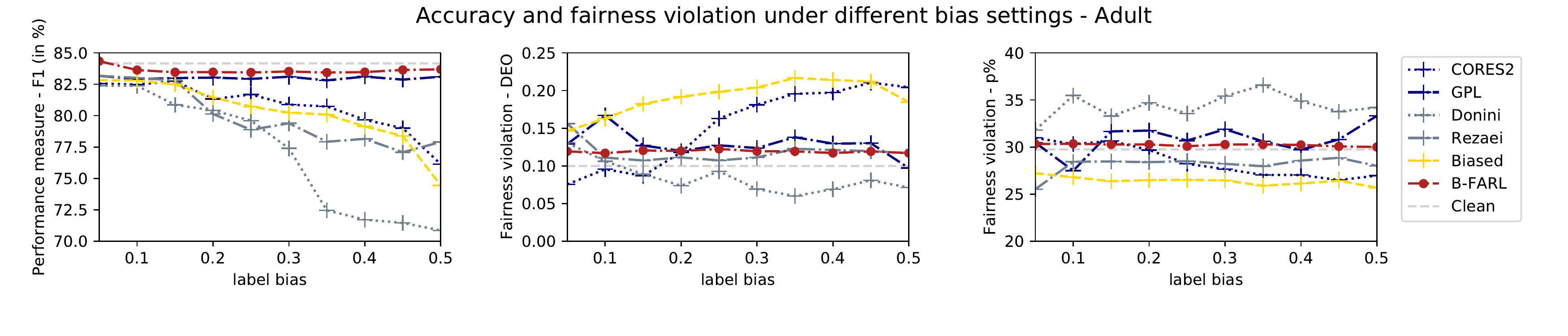}
\includegraphics[width=1\textwidth]{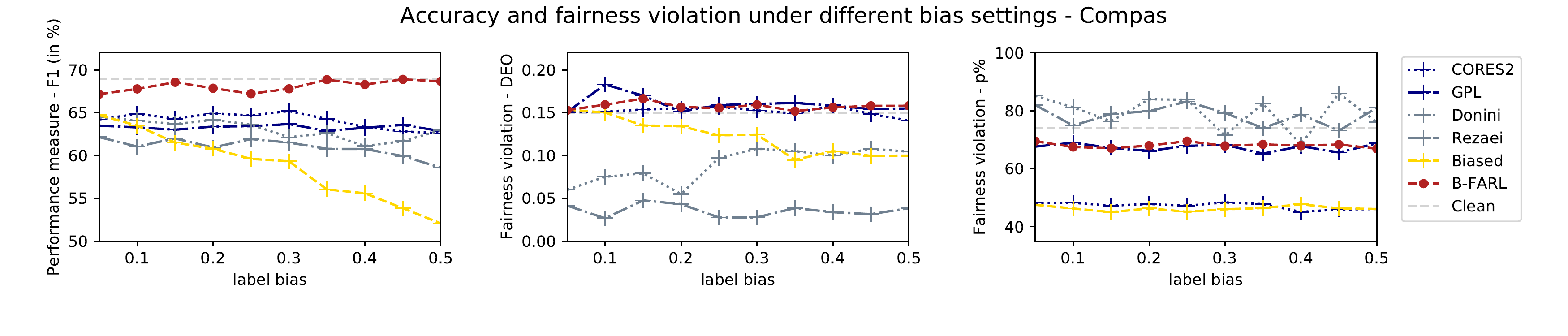}
\includegraphics[width=1\textwidth]{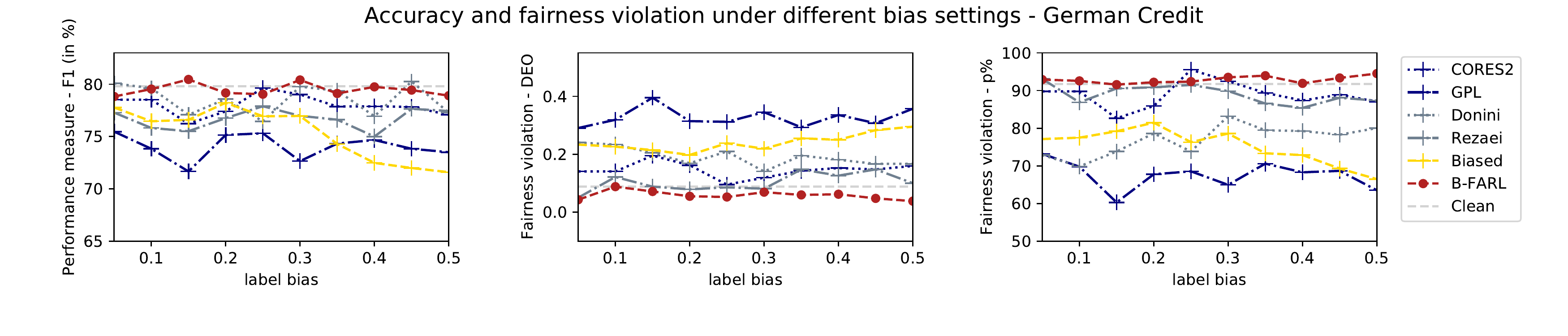}
\end{center}
\caption{Accuracy and fairness violation under different label bias settings. The x-axis is the average label bias over $\{\theta_0^+,\theta_0^-,\theta_1^+, \theta_1^-\}$. We use same color to denote the methods in the same category, i.e., we use blue color to denote GPL and $\text{CORES}^2$, which are both noisy label learning method, and we use gray color to denote two algorithmic fairness methods.} 
\label{fig:label_bias}
\end{figure}

\subsection{Comparison and Application on Real Word Data}
\subsubsection{Case 1: Label bias}
\label{sec:case_1}
In the first case, we test the performance of different methods under different settings of label bias with selection bias fixed. We set average label bias amount from 0.1 to 0.5 while fix the selection bias with $\sigma=1.1$. We add bias into the train set only while keep test set clean. In the settings, we always require $\theta_0^+ > \theta_0^-$ and $\theta_1^- > \theta_1^+$.

The results are shown in Figure.~\ref{fig:label_bias}. The prediction performance of our method generally outperforms other methods with the increase of label bias. Overall, the two algorithmic fairness methods have lower F1 scores than the two noisy label learning methods and B-FARL, though they have lower fairness violations. This demonstrates the algorithmic fairness methods will achieve a certain fairness level by ``flipping" the labels of some individuals, and the low F1 indicates the flipping is in the opposite direction of the true labels. This is what we have claimed the controversy of accuracy and fairness trade-off. Also, we notice that the F1 score of two algorithmic fairness methods decreases while the fairness violation increases as the amount of label bias increases, which indicates they are not robust to the different amount of label bais; In the meantime, two noisy label learning methods, as well as B-FARL, have more steady F1 when we increase the amount of label bias. However, since $\text{CORES}^2$ does not take fairness into consideration, it has an overall higher fairness violation compared to GPL and B-FARL. GPL deploys derived fairness constraints under corrupted distribution, so it has overall lower fairness violation compared to $\text{CORES}^2$, but higher than B-FARL. 

For the adult dataset, we found the results for GPL are very close to ours while GPL has a slightly higher p\% value and DEO, and ours has higher accuracy and lower DEO. For the Compas dataset, the accuracy of our method is closest to the accuracy on the clean data and achieves closer p\% to the benchmark for clean distribution. For the German Credit dataset, B-FARL has the highest f1, with the highest p\% and lowest DEO. Overall, B-FARL is superior to the other baseline methods for optimizing towards the latent fair labels under different label bias amounts. 

\subsubsection{Case 2: Selection Bias}
In this section, we conduct our experiments on how selection bias would affect performance and fairness violation. We fixed the label bias which we set as $\theta_0^+ = 0.25$, $\theta_0^- = 0.05$, $\theta_1^+ = 0.05$ and $\theta_1^- = 0.25$. We increase the selection bias by 2\% from $\sigma=1.01$ to $\sigma=1.1$. Similar to the setting in Sec~\ref{sec:case_1}, we add selection bias to train set only.

\begin{figure}[!htp]
\begin{center}
\includegraphics[width=1.0\textwidth]{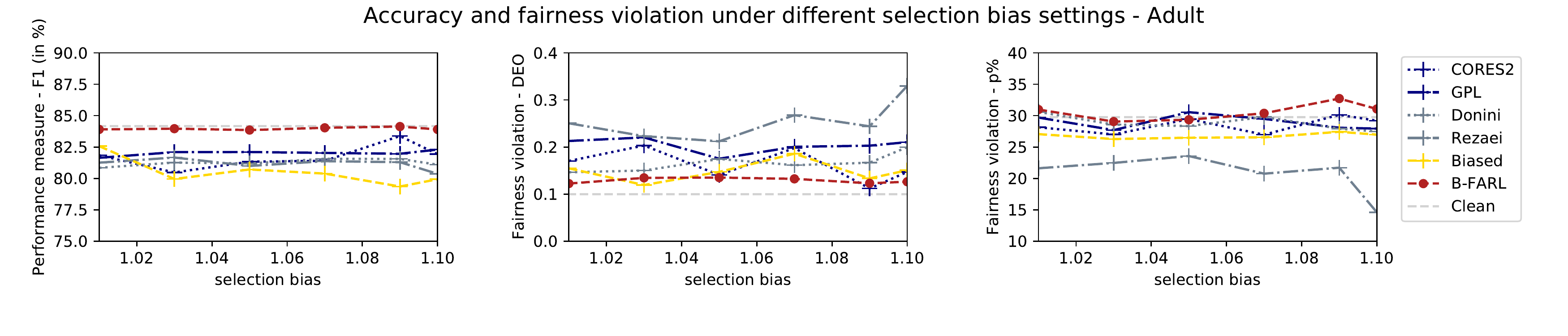}
\includegraphics[width=1.0\textwidth]{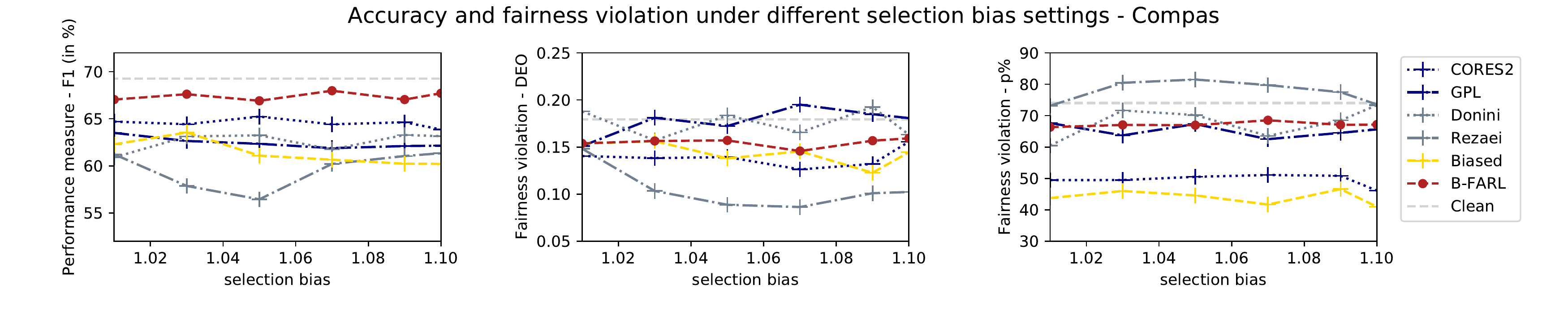}
\includegraphics[width=1.0\textwidth]{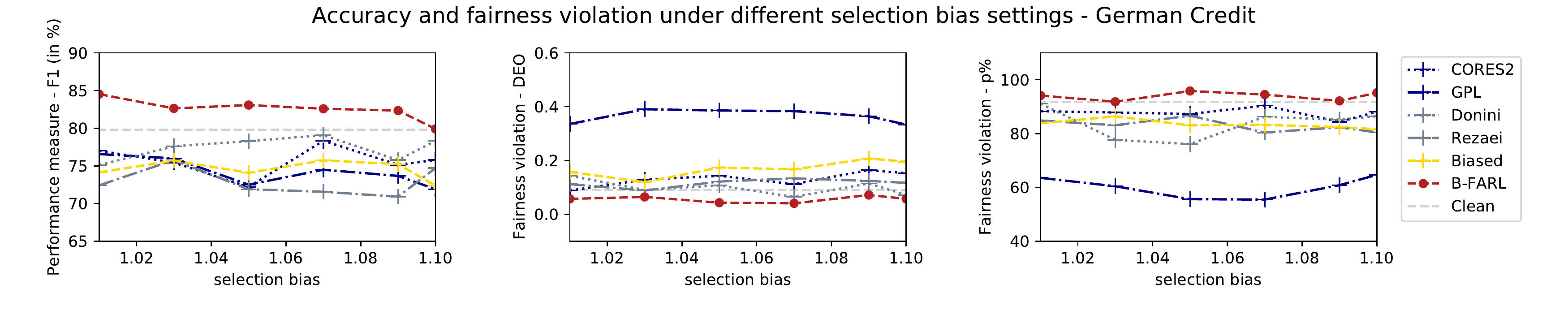}
\end{center}
\caption{Accuracy and fairness violation under different selection bias settings. The x-axis is the average selection bias which is related to the proportion of positive labeled instances among the protected group. The blue color is for GPL and $\text{CORES}^2$, which are both noisy label learning method. The gray is for two algorithmic fairness methods.}\label{fig:selection_bias}
\end{figure}

From Fig.~\ref{fig:selection_bias} we can see B-FARL also outperforms among all the methods with the highest F1 and low fairness violations. Unlike the experimental results of label bias, we do not observe an apparent decreasing trend as selection bias increases. However, the difference between our method and other methods are distinct. And our performance is the closest to the clean one. Also, we found GPL cannot handle selection bias very well compared to its performance under label bias. For the Adult dataset, B-FARL has the highest F1 and lowest fairness violation w.r.t. both DEO and p\% measure and is close to the baseline on clean data. The F1 score of two algorithmic fairness methods and two noisy label learning methods are close. For the Compas and German Credit dataset, B-FARL has the highest F1 score. Two algorithmic fairness methods have the highest p\% value. Still, the method proposed by ~\cite{donini2020empirical} has a higher DEO violation and higher F1 than the method proposed by ~\cite{Rezaei_2020}. In contrast, the method proposed by ~\cite{Rezaei_2020} has the lowest F1 and lowest DEO violation. This demonstrates the same phenomenon we have concluded in Sec~\ref{sec:case_1}. Similar to the experiment of label bias, the two noisy label learning methods have higher F1 and higher fairness violations compared to the two algorithmic fairness methods. B-FARL has the highest F1 and lowest fairness violation compared to all the methods. Overall, B-FARL is superior to the other baseline methods also under different amounts of selection bias.

\subsection{Evaluate Our Methods on the Clean Data}
We also evaluate our method on the clean data directly. We simulated ten sets of clean data according to Fig.~\ref{fig:generated}. The detailed generation steps are provided in Appendix~\ref{supp:synthetic_generating}. We found our method can achieve similar accuracy and fairness level to the baseline on the clean data. Though GPL has the highest F1 score, it also has the highest fairness violations, this may imply GPL over-corrects the labels. In contrast, ~\cite{Rezaei_2020} has the smallest fairness violations but with lowest F1 score, this was aligned with the results in Section~\ref{sec:case_1}. We found both $\text{CORES}^2$ and ~\cite{donini2020empirical} have accuracy and fairness drop, the former may due to the nonlinear measure of fairness constraints, which has the adverse impact of both performance and fairness, the latter may caused by the second phase of sample sieve, which introduce randomness for the semi-supervised learning.  
\begin{table}[!ht]
\centering
\resizebox{\linewidth}{!}{%
\begin{tabular}{|c||c|c|c|c|c|c|} 
\toprule
 & Clean & B-FARL & Donini & Rezaei & $\text{CORES}^2$ & GPL \\
 \hline
F1 & 98.52$\pm$1.28\% & 98.51$\pm$1.60\% & 98.22$\pm$0.85\% & 94.93$\pm$0.89\% & 94.86$\pm$2.23\% & 98.95$\pm$0.66\%\\

DEO &0.62$\pm$0.61\% & 0.71$\pm$0.73\% & 0.79$\pm$0.34\% & 0.46$\pm$0.40\% & 0.87$\pm$0.47\% & 1.06$\pm$0.77\%\\

p\% & 95.10$\pm$4.14\% & 95.39$\pm$3.80\% & 94.26$\pm$3.06\% &95.88$\pm$4.32\% & 95.05$\pm$4.13\% & 94.41$\pm$4.47\%\\
\bottomrule
\end{tabular}
}
\caption{Performance on the clean datasets}
\label{tab:clean_data}
\end{table}

\subsection{Impact of Regularization Intensity}

\begin{wrapfigure}{R}{0.35\textwidth}
\vspace{-.2in}
\hspace{-.2in}
   \centering
        \includegraphics[width=0.25\textwidth]{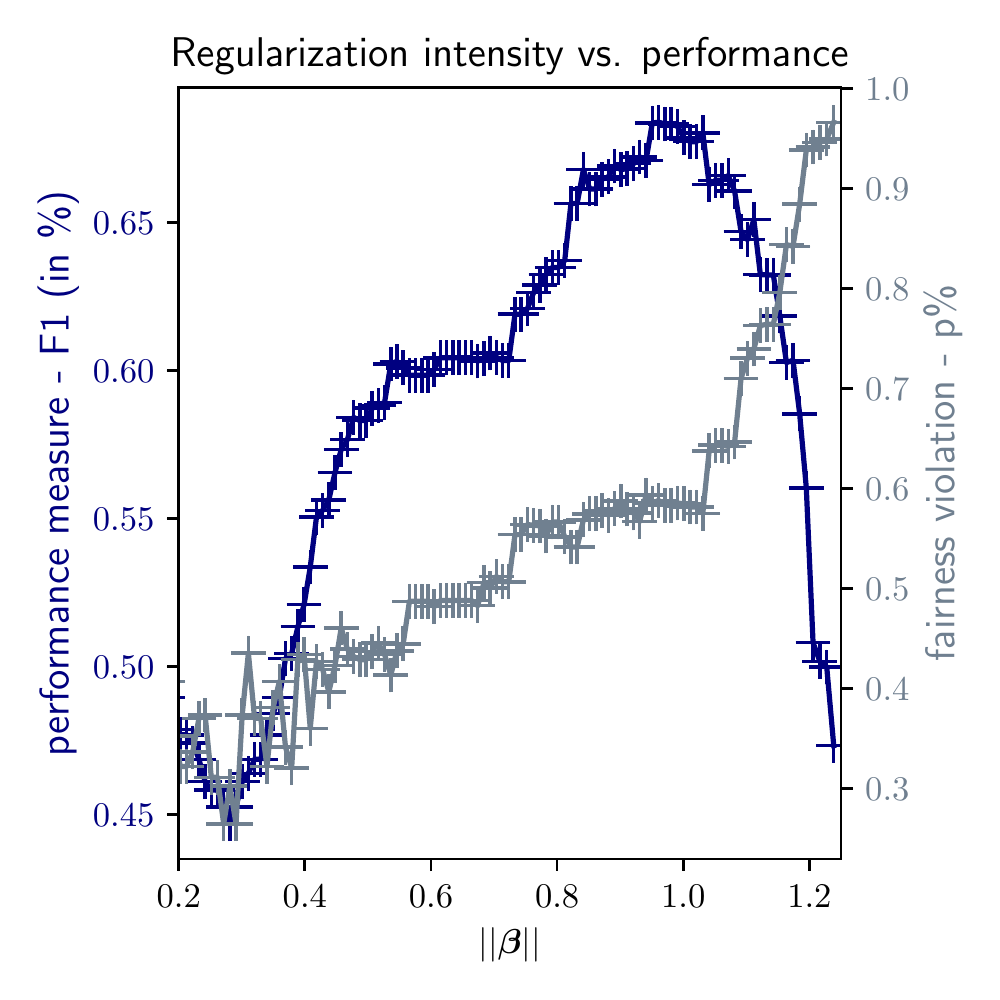}
 \vspace{-.2in}
   \caption{Regularization intensity vs. performance on Compas dataset.}
   \label{fig:reg}
 \end{wrapfigure}
We also examine how the regularization intensity $\bm{\beta}$ works by conducting the experiment on the `Compas' dataset. We record the F1 score and p\% value when increasingly update $\bm{\beta}$. We compute $||\bm{\beta}||$ to measure the intensity. We can see from Fig.~\ref{fig:reg}, when the regularization intensity increases from around 0.2 to 0.95, the performance and p\% value also increases. This demonstrates that when B-FARL is guided by appropriate regularization intensity, the accuracy and fairness improve simultaneously. However, as the intensity gets larger, we can see the p\% value still increases, but the F1 score starts to decrease. This indicates that the fairness regularizer term starts to dominate as the intensity becomes larger and hence causes the results to achieve perfect fairness while neglecting the accuracy performance. However, with appropriate regularization intensity, the accuracy performance and fairness improve together.



\section{Conclusion}
\label{conclusion}
In this paper, we tackle the discrimination issue from the label bias and selection bias perspective. We propose a bias-tolerant fair classification method by designing B-FARL, which is a loss having the regularization effect that can compensate both label bias and selection bias. To optimize B-FARL more efficiently, we incorporate it with the model-agnostic meta-learning framework to update the hyperparameters. Besides, We decompose the B-FARL loss into three meaningful components, including expected loss under the distribution of clean samples, fairness regularizer, and a regularizer on the disagreement between biased and unbiased observations to demonstrate the effectiveness of B-FARL theoretically. We empirically demonstrated the superiority of our proposed framework through experiments. A future research direction of this work is to relax the assumption that $X$ is independent of $A$ for more complex data since in practice $X$ will always contain the information from $A$. This can also be connected with instance-dependent label bias setting since we do not only consider the flip rate related to the true label and $A$, but rather include the dependency with $X$.

\bibliographystyle{plain}
\bibliography{references}

\newpage
\appendix
\onecolumn

\section{Decompose the loss}\label{apd:decompose}

We decompose the new loss:
\begin{equation}
    \mathbb{E}_{\widetilde{D}}[\ell_B(\bm{\omega}) + \bm{\beta}\bm{\ell}_A(\bm{\omega})]
\end{equation}
For simplicity, we omit $\bm{\omega}$ from $\ell{f(x,\bm{\omega})}$ in the following derivations. So we first decompose the first term: $\mathbb{E}_{\widetilde{D}}[\ell(f(X),Y)$

{\small
\begin{equation}
\label{eq:part1}
    \begin{aligned}
    &\mathbb{E}_{\widetilde{D}}[\ell_B(\bm{\omega})] \\ 
    & = \sum_{k\in [C]}\sum_{l \in [C]}\sum_a \int_x P(Y=k\mid Z=l,A=a,x)P(x\mid Z=l,A=a) \delta_a \ell (f(x),k) dx P(Z=l)P(A=a)\\
    & = \sum_{k\in [C]}\sum_{l \in [C]}\sum_a P(Z=l)P(A=a) \mathbb{E}_{D_{x \mid Z=l,A=a}}[P(Y=k\mid Z=l,A=a)\delta_a \ell(f(x),k)] \\
    & = \sum_{k\in [C]}\sum_{l \in [C]}\sum_a P(Z=l)P(A=a)[\underbrace{\mathbb{E}_{D_{x \mid Z=l,A=a}} P(Y=k\mid Z=l,A=a)\cdot \mathbb{E}_{D_{x \mid Z=l,A=a}} \delta_a \ell(f(x),k)}_A\\
    &+\underbrace{\text{Cov}_{D_{x \mid Z=l,A=a}} [P(Y=k\mid Z=l,A=a)\cdot \delta_a \ell(f(x),k)}_B]
    \end{aligned}
\end{equation}}
First expand Part A in \cref{eq:part1}, we can obtain:
\small{
\begin{equation}
\label{eq:17}
    \begin{aligned}
     &\sum_a P(A=a) \delta_a \sum_{k\in [C]} [P(Z=k)\cdot \mathbb{E}_{D_{x \mid Z=k,A=a}} P(Y=k\mid Z=k,A=a)\cdot \mathbb{E}_{D_{x \mid Z=k,A=a}} \ell(f(x),k)\\
    &+ \sum_{l\in[C],l\neq k}P(Z=l)\cdot \mathbb{E}_{D_{x \mid Z=l,A=a}} P(Y=k\mid Z=l,A=a)\cdot \mathbb{E}_{D_{x \mid Z=l,A=a}} \ell(f(x),k) ] \\
    & = \sum_a P(A=a)\delta_a [ P(Z=1)\cdot \mathbb{E}_{D_{x\mid Z=1,A=a}}(1-\theta_a^{-})\cdot \mathbb{E}_{D_{x\mid Z=1,A=a}}\ell(f(x),1)\\
    &+P(Z=-1)\cdot \mathbb{E}_{D_{x\mid Z=-1,A=a}}(1-\theta_a^{+})\cdot \mathbb{E}_{D_{x\mid Z=-1,A=a}}\ell(f(x),-1)]\\
    &+ \sum_a P(A=a) \delta_a \sum_{k\in [C]} \sum_{l\in[C],l\neq k}P(Z=l)\cdot \mathbb{E}_{D_{x \mid Z=l,A=a}} P(Y=k\mid Z=l,A=a)\cdot \mathbb{E}_{D_{x \mid Z=l,A=a}} \ell(f(x),k) ]  \\
    &=\sum_a P(A=a) \delta_a [ \underbrace{\mathbb{E}_{D_{x}}(1-\theta_a^{-} - \theta_a^{+})\cdot \mathbb{E}_{D\mid a}\ell (f(x),Z))}_C\\
    &+ \underbrace{P(Z=1)\cdot \mathbb{E}_{D_{x\mid Z=1,A=a}}\theta_a^{+}\cdot \mathbb{E}_{D_{x\mid Z=1,A=a}}\ell(f(x),1)+ P(Z=-1)\cdot \mathbb{E}_{D_{x\mid Z=-1,A=a}}\theta_a^{-}\cdot \mathbb{E}_{D_{x\mid Z=-1,A=a}}\ell(f(x),-1)]}_D\\
     &+ \underbrace{\sum_a P(A=a) \delta_a \sum_{k\in [C]} \sum_{l\in[C],l\neq k}P(Z=l)\cdot \mathbb{E}_{D_{x \mid Z=l,A=a}} P(Y=k\mid Z=l,A=a)\cdot \mathbb{E}_{D_{x \mid Z=l,A=a}} \ell(f(x),k) }_E
    \end{aligned}
\end{equation}}
Now let's expand part B in \cref{eq:part1}, we can get:
{\small
\begin{equation}
\label{eq:part2}
    \begin{aligned}
    & \sum_{k\in [C]}\sum_{l \in [C]}\sum_a P(Z=l)P(A=a) \text{Cov}_{D_{x \mid Z=l,A=a}} [P(Y=k\mid Z=l, A=a)\cdot \delta_a \ell(f(x),k)] \\
    &=  \sum_a P(A=a) \delta_a [\sum_{k\in [C]} P(Z=k) \mathbb{E}_{D_{x \mid Z=k,A=a}}((P(Y=k\mid Z=k,A=a)-\mathbb{E}_{D_{x \mid Z=k,A=a}}(P(Y=k\mid Z=k,A=a))\\
    &\times (\ell (f(x),k)- \mathbb{E}_{D_{x \mid Z=k,A=a}}[\ell (f(x),k])
    \\ &+  \sum_{k\in [C]} \sum_{l \in [C], l\neq k} P(Z=l) \mathbb{E}_{D_{x \mid Z=l,A=a}}((P(Y=k\mid Z=l,A=a)-\mathbb{E}_{D_{x \mid Z=l,A=a}}(P(Y=k\mid Z=l,A=a))\\
    &\times (\ell (f(x),k)- \mathbb{E}_{D_{x \mid Z=l,A=a}}[\ell (f(x),k])]
    \end{aligned}
\end{equation}}
If we combine \cref{eq:part2} with Part E in \cref{eq:17}, we can obtain:
{\small
\begin{equation}
\label{eq:part3}
    \begin{aligned}
   &\sum_a P(A=a) \delta_a \sum_{k \in [C]}[\sum_{l\in [C],l\neq k}P(Z=l) \mathbb{E}_{D_{x \mid Z=l,A=a}}(P(Y=k\mid Z=l,A=a)\ell (f(x),k)\\
    &+ P(Z=k)\mathbb{E}_{D_{x \mid Z=k,A=a}}((P(Y=k\mid Z=k,A=a)-\mathbb{E}_{D_{x \mid Z=k,A=a}}(P(Y=k\mid Z=k,A=a))[\ell (f(x),k)]]\\
    &= \sum_a P(A=a) \delta_a [P(Z=1)\mathbb{E}_{D_{x \mid Z=1,A=a}}(1-\theta_a^{-}-\mathbb{E}_{D_{x \mid Z=1,A=a}}(1-\theta_a^-))\ell(f(x),1)\\
    &+P(Z=-1)\mathbb{E}_{D_{x \mid Z=-1,A=a}}(1-\theta_a^{+}-\mathbb{E}_{D_{x \mid Z=-1,A=a}}(1-\theta_a^+))\ell(f(x),-1)\\
    &+P(Z=-1)\mathbb{E}_{D_{x \mid Z=-1,A=a}}(\theta_a^{+}\ell(f(x),1)]+P(Z=1)\mathbb{E}_{D_{x \mid Z=1,A=a}}(\theta_a^{-}\ell(f(x),-1)]
    \end{aligned}
\end{equation}}
Finally, we combine \cref{eq:part3} with part C as well as part D in \cref{eq:17} and we can finally get the decomposed terms:
\begin{equation}
\label{eq:decompose_first_part}
    \begin{aligned}
    &\mathbb{E}_{\widetilde{D}}[\ell_B(\bm{\omega})]\\
    &=\sum_a P(A=a) \delta_a [(1-\theta_a^+ - \theta_a^-)\mathbb{E}_{D\mid A=a}\ell(f(x),Z)+\sum_{k \in [C]}\sum_{l\in [C]}P(Z=l)\mathbb{E}_{D_{x\mid Z=l,A=a}}U_{lk}\ell(f(x),k)]\\
    &=\sum_a P(A=a) [\mathbb{E}_{D\mid A=a}\ell(f(x),Z)+\sum_{k \in [C]}\sum_{l\in [C]}P(Z=l)\mathbb{E}_{D_{x\mid Z=l,A=a}}  U_{lk}\ell(f(x),k)]\\
    &=\mathbb{E}_{D}[\ell(f(X),Z)]+\sum_a P(A=a) \sum_{k \in [C]}\sum_{l\in [C]}P(Z=l)\mathbb{E}_{D_{x\mid Z=l,A=a}} U_{lk}\ell(f(x),k)
    \end{aligned}
\end{equation}
where
 \begin{equation*}
  U_{lk}(x,a) = \begin{cases}
        \delta_a \theta_a^{sgn(k)} \text{     if $l \neq k$,}\\
        \delta_a \theta_a^{sgn(l)} \text{    if $l=k$}.
        \end{cases}
 \end{equation*}
Now we then decompose the second and third term in \cref{eq:new_loss}.
{\small 
\begin{equation}
\label{eq:part5}
\begin{aligned}
   & \mathbb{E}_{\widetilde{D}} [\bm{\beta} \bm{\ell}_{A}(\bm{\omega})] \\
   &= \mathbb{E}_{\widetilde{D}}[- \beta_0 \cdot \mathbb{E}_{Y\mid  \widetilde{D},A=0} (1-a_i)\ell(f(x),Y) - \beta_1 \cdot \mathbb{E}_{Y\mid \widetilde{D},A=1}a_i \ell(f(x),Y)]\\
   &=\mathbb{E}_{\widetilde{D}}[\lambda \cdot (\mathbb{E}_{Y\mid  \widetilde{D},A=0} (1-a_i)\ell(f(x),Y) - \mathbb{E}_{Y\mid \widetilde{D},A=1}a_i \ell(f(x),Y)) \\
   &- \rho_a \cdot \mathbb{E}_{Y\mid  \widetilde{D},A=0} (1-a_i)\ell(f(x),Y) - \rho_b \cdot\mathbb{E}_{Y\mid \widetilde{D},A=1}a_i \ell(f(x),Y)] \\
   & = \lambda \cdot [\mathbb{E}_{\widetilde{D}\mid A=0}\ell(f(x),Y) - \mathbb{E}_{\widetilde{D}\mid A=1}\ell_{A=1}(f(x),Y)]
   - \rho_a(\int_x\sum_k P(X=x,Y=k,A=0)(1-0)\ell(f(x),k)dx\\
   &+ \int_x\sum_k P(X=x,Y=k,A=1)(1-1)\ell(f(x),k)dx)\\ 
   &- \rho_b(\int_x\sum_k P(X=x,Y=k,A=0)(0)\ell(f(x),k)dx + \int_x\sum_k P(X=x,Y=k,A=1)(1)\ell(f(x),k)dx)\\ 
   & = \lambda \cdot [\mathbb{E}_{\widetilde{D}\mid A=0}\ell(f(x),Y) - \mathbb{E}_{\widetilde{D}\mid A=1}\ell_{A=1}(f(x),Y)]-\rho_a\cdot \int_x\sum_k P(X=x,Y=k,A=0)(1-0)\ell(f(x),k)dx \\
   &- \rho_b \int_x\sum_k P(X=x,Y=k,A=1)(1)\ell(f(x),k)dx \lambda \cdot \mathbb{E}_{\widetilde{D}}[\ell_{A=0}(f(x),Y) - \ell_{A=1}(f(x),Y)) \\&- P(A=0)\rho_a\cdot \sum_k\sum_l P(Z=l)\mathbb{E}_{D_{x \mid l,0}}P(Y=k)\ell(f(x),k)\\
   &-P(A=1)\rho_b\cdot \sum_k\sum_l P(Z=l)\mathbb{E}_{D_{x \mid l,1}}P(Y=k)\ell(f(x),k)\\
   &= \lambda \cdot [\mathbb{E}_{\widetilde{D}\mid A=0}\ell(f(x),Y) - \mathbb{E}_{\widetilde{D}\mid A=1}\ell_{A=1}(f(x),Y)] - \sum_aP(A=a)\sum_k\sum_l P(Z=l)\mathbb{E}_{D_{x \mid l,a}}\gamma_a\cdot P(Y=k)\ell(f(x),k)
\end{aligned}
\end{equation}}
where $\lambda = \rho_a - \beta_0 = \beta_1 -\rho_b$ and $\gamma_a = \begin{cases} \rho_a \text{ if } a=0\\ \rho_b \text{ if } a=1 \end{cases}$. We can also decompose the second term and third term in this way
{\small 
\begin{equation}
\label{eq:part6}
\begin{aligned}
   & \mathbb{E}_{\widetilde{D}} [\bm{\beta} \bm{\ell}_{A}(\bm{\omega})] \\
   &=\mathbb{E}_{\widetilde{D}}[\lambda \cdot (\mathbb{E}_{Y\mid \widetilde{D},A=1}a_i \ell(f(x),Y)-\mathbb{E}_{Y\mid  \widetilde{D},A=0} (1-a_i)\ell(f(x),Y) ) \\
   &- \rho_b \cdot \mathbb{E}_{Y\mid  \widetilde{D},A=1} a_i\ell(f(x),Y) - \rho_a \cdot\mathbb{E}_{Y\mid \widetilde{D},A=0}(1-a_i) \ell(f(x),Y)] \\
   &= \lambda \cdot [\mathbb{E}_{\widetilde{D}\mid A=1}\ell(f(x),Y) - \mathbb{E}_{\widetilde{D}\mid A=0}\ell(f(x),Y)] - \sum_aP(A=a)\sum_k\sum_l P(Z=l)\mathbb{E}_{D_{x \mid l,a}}\gamma_a\cdot P(Y=k)\ell(f(x),k)
\end{aligned}
\end{equation}}
where $\lambda = \beta_0 - \rho_a = \rho_b-\beta_1$. Then by combining \cref{eq:part5} and \cref{eq:part6} we can get:
{\small 
\begin{equation}
\label{eq:part7}
\begin{aligned}
   & \mathbb{E}_{\widetilde{D}} [\bm{\beta} \bm{\ell}_{A}(\bm{\omega})] \\
   &= \lambda \cdot |\mathbb{E}_{\widetilde{D}\mid A=0}\ell(f(x),Y) - \mathbb{E}_{\widetilde{D}\mid A=1}\ell(f(x),Y)| - \sum_aP(A=a)\sum_k\sum_l P(Z=l)\mathbb{E}_{D_{x \mid l,a}}\gamma_a\cdot P(Y=k)\ell(f(x),k)
\end{aligned}
\end{equation}}
where $\lambda = |\beta_0 - \rho_a| = |\rho_b-\beta_1|$. Finally, we combine \cref{eq:decompose_first_part} and \cref{eq:part7} together and get:

{\small 
\begin{equation}
    \begin{aligned}
    & \mathbb{E}_{\widetilde{D}}[\ell_B(\bm{\omega}) + \bm{\beta} \bm{\ell}_{A}(\bm{\omega})]\\
    &=\mathbb{E}_{D}\ell(f(X),Z)] + \sum_a P(A=a)\sum_{k \in [C]}\sum_{l\in [C]}P(Z=l)\mathbb{E}_{D_{x \mid l,a}}(U_{lk}(x,a)-\gamma_a\cdot P(Y=k))\ell (f(x),k)]\\& + \lambda \cdot |\mathbb{E}_{\widetilde{D}\mid A=1}\ell(f(x),Y) - \mathbb{E}_{\widetilde{D}\mid A=0}\ell(f(x),Y)|
    \end{aligned}
\end{equation}}


\section{Derive the relationship between selection bias and label bias}\label{apd:second}
Let $\widetilde{N}_{sign(y),a}$, $\hat{N}_{sign(y),a}$ and $N_{sign(y),a}$ denote the number of instances in group with membership of $(sign(y),a)$. Here $\widetilde{N}_.$ is for the observed data with both biases. $\hat{N}_.$ is for the data with selection bias only. 

\begin{equation}
    \widetilde{N}_{+1,1} = (1-\theta_1^-)\cdot \hat{N}_{+1,1} + \theta_1^+\cdot \hat{N}_{-1,1}
\end{equation}

Let $\varepsilon_0^-$ denotes the bias rate combining the selection bias and label bias.

\begin{equation}
    \widetilde{N}_{+1,1} = (1-\varepsilon_1^-)\cdot N_{+1,1} + \varepsilon_1^+\cdot N_{-1,1}
\end{equation}

We assume the selection bias is proportion to the ratio of positive labeled instances in unprotected group, i.e.,
\begin{equation}
\begin{aligned}
    \frac{\hat{N}_{+1,1}}{\hat{N}_{+1,1}+N_{-1,1}} &= \frac{r}{\sigma} = \frac{N_{+1,1}}{\sigma(N_{+1,1}+N_{-1,1})} \\
    \hat{N}_{+1,1} &= \frac{1-r}{\sigma-r}N_{+1,1}
\end{aligned}
\end{equation}

Then we can derive the relationship between $\varepsilon_1^+$ and $\theta_1^+$ by 
\begin{equation}
    \begin{aligned}
    (1-\varepsilon_1^-)\cdot N_{+1,1} + \varepsilon_1^+\cdot N_{-1,1} &= (1-\theta_1^-)\cdot \hat{N}_{+1,1} + \theta_1^+\cdot \hat{N}_{-1,1}\\
    (1-\theta_1^+)\frac{1-r}{\sigma-r}N_{+1,1} &= (1-\varepsilon_1^-)N_{+1,1}\\
    \theta_1^- &= \frac{\sigma-r}{1-r}\varepsilon_1^- + \frac{1-\sigma}{1-r}
    \end{aligned}
\end{equation}

\section{Synthetic data generating process}
\label{supp:synthetic_generating}
\begin{itemize}
    \item Generate $W \sim N(0,\sigma)$ (we use $\sigma = I^{15\times 15}$, and dimension of $W$ is 15).\\
    \item Generate $a_i \sim \text{Bernoulli}(\alpha)$, (we set $\alpha$ = 0.1 and n = 2000).
    \item Generate $x_i^j \sim \text{Bernoulli}(\frac{1}{j+1}^r)$ for $j = 0,...,k-2$, where $k$ is the dimension of $W$, which is 15. $r$ controls the discrepancy between the rarity of features. We sample each dimension $i$ according to a Bernoulli proportional to $\frac{1}{i}$ making some dimensions common and others rare (we set $r = 0.5$). \\
    \item Generate unbiasd label $z_i = \text{max}(0,\text{sign}(w^T_{\text{gen}}x_i))$ \\
    \item Generate biased label $y_i \sim g(y\mid z_i,a_i,x_i,\beta)$ \\
    where $g(y_i \mid z_i,a_i,x_i,\beta)  = \begin{cases} \beta \text{ if } y_i \neq z_i \wedge z = a_i \\ 1-\beta \end{cases}$ and $\beta$ controls the amount of label bias (We set $\beta$ = 0.5). 
\end{itemize}

\end{document}